# Automatic Construction of Chinese Herbal Prescriptions From Tongue Images Using CNNs and Auxiliary Latent Therapy Topics

Yang Hu, Guihua Wen, Huiqiang Liao, Changjun Wang, Dan Dai, and Zhiwen Yu, *Senior Member, IEEE*

*Abstract*—The tongue image provides important physical information of humans. It is of great importance for diagnoses and treatments in clinical medicine. Herbal prescriptions are simple, noninvasive, and have low side effects. Thus, they are widely applied in China. Studies on the automatic construction technology of herbal prescriptions based on tongue images have great significance for deep learning to explore the relevance of tongue images for herbal prescriptions, it can be applied to healthcare services in mobile medical systems. In order to adapt to the tongue image in a variety of photographic environments and construct herbal prescriptions, a neural network framework for prescription construction is designed. It includes single/double convolution channels and fully connected layers. Furthermore, it proposes the auxiliary therapy topic loss mechanism to model the therapy of Chinese doctors and alleviate the interference of sparse output labels on the diversity of results. The experiment use the real-world tongue images and the corresponding prescriptions and the results can generate prescriptions that are close to the real samples, which verifies the feasibility of the proposed method for the automatic construction of herbal prescriptions from tongue images. Also, it provides a reference for automatic herbal prescription construction from more physical information.

*Index Terms*—Convolution channels, neural networks, prescriptions construction, therapy topics, tongue images.

## I. INTRODUCTION

CHINA has a vast and increasing population. Making full use of medical resources to meet the needs of the patients is always an urgent concern [1]. Healthcare informatics [2], [3], especially the rapid development of mobile medical services, can help immobile and mobile people obtain cheap and high-quality medical services [4]–[6].

In China, herbal prescriptions contain low side effects and provide a simple, cheap, natural, and noninvasive treatment or health preservation option [7]. The tongue image is a potentially important body sign in clinical diagnosis and treatment. Therefore, there have already been extensive studies on the collection [8], [9], processing [10], [11] and analysis [12], [13] of tongue images, which can assist in diagnosis and treatment. Zhang *et al.* [14] integrated the color, texture, and geometric features of tongue images to detect diabetes mellitus and nonproliferative diabetic retinopathy. Li *et al.* [15] combined the color, texture, and geometric features of the tongue image, facial parts, and sublingual information to detect diabetes and early glycemia. However, previous studies usually require a professional tongue image acquisition device, these devices usually do not appear in some community hospitals with poor conditions. The previous studies also lack robustness to the requirements of the camera, shooting angle, illumination, and other conditions, which is not conducive to expand on mobile terminals to convenient for carrying. Therefore, the research on the feature coding of low-quality tongue images and the auxiliary construction of herbal prescriptions has research significance and application value.

Deep learning based on neural networks has been widely applied in medical informatics [16], [17], especially in medical image processing [18]–[21] in recent years. It covers medical image preprocessing [22], [23], medical image classification [24], [25] and other directions and provides great technical support for auxiliary diagnoses and treatments. Tang *et al.* [26] encoded the multiscale medical image features using sparse autoencoders with different receptive field sizes and generated the feature maps for classification using the convolution operation. Zhang *et al.* [27] enhanced the multiscale feature ensembles and utilization efficiency and integrated the improved attention mechanism to read the images and generate diagnostic reports.

Also, artificial intelligence plays an important role in medicine recommendations [28], medical simulation [29], and clinical decision support [30]. Hoang *et al.* [31] used social media as the information source to mine sequences of drugs and adverse effects that signal detrimental prescription cascades. Long and Yuan [34] described an accurate,

Manuscript received January 22, 2019; revised March 30, 2019; accepted April 3, 2019. This work was supported in part by the Guangdong Province Higher Vocational Colleges and Schools Pearl River Scholar Funded Scheme under Grant 2018, in part by the National Natural Science Foundation of China under Grant 60973083, Grant 61273363, Grant 61722205, Grant 61751205, Grant 61751202, Grant 61572199, Grant 61502174, and Grant U1611461, in part by the Science and Technology Planning Projects of Guangdong Province under Grant 2014A010103009, Grant 2015A020217002, and Grant 2018B010107002, and in part by the Guangzhou Science and Technology Planning Project under Grant 201504291154480. This paper was recommended by Associate Editor Y. Zhang. *(Corresponding author: Guihua Wen.)*

Y. Hu, G. Wen, H. Liao, D. Dai, and Z. Yu are with the School of Computer Science and Engineering, South China University of Technology, Guangzhou 510000, China (e-mail: superhy199148@hotmail.com; crghwen@scut.edu.cn).

C. Wang is with Department of Chinese Medicine, Guangdong General Hospital, Guangzhou 510000, China.

This paper has supplementary downloadable material available at http://ieeexplore.ieee.org, provided by the author.

Color versions of one or more of the figures in this paper are available online at http://ieeexplore.ieee.org.

Digital Object Identifier 10.1109/TCYB.2019.2909925

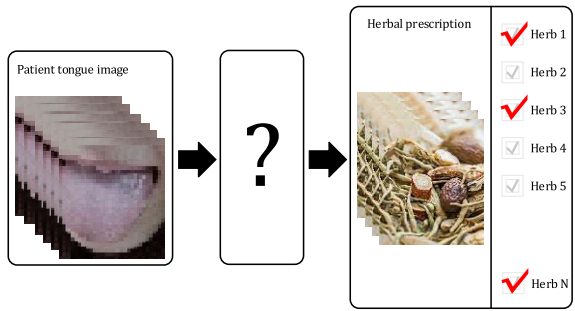

Fig. 1. Neural network herbal prescription construction task.

computationally efficient, and scalable algorithm to construct the medication history timeline that can inform clinical decision making. Yu *et al.* pioneered a traditional Chinese medicine (TCM) knowledge graph that integrated terms, databases, and other resources. It provides services, such as visualization, knowledge retrieval, and recommendations; thus, this system has great significance to the studies of medical clinical decision support in China [35].

In this paper, we intend to use convolutional networks to model the relationship between tongue images and Chinese herbal prescriptions. We also introduce multitask structure and amalgamate the main and auxiliary features with features fusion strategy, as in [32] and [33].

Furthermore, we develop an assistant prescription system based on this paper, which is also applied in several basic community hospitals. It provides tips and references for interns or young doctors in prescribing, to improve their efficiency. The final prescriptions is decided by doctors.

To the best of our knowledge, there has not been any work using tongue images to construct herbal prescription. In order to robustly model the tongue images of patients in various photographic environments and accurately excavate the association between tongue information and herbal prescriptions (as shown in Fig. 1), the following main work is carried out in this paper.

1) A neural framework of single/double convolution channels and multiple fully connected layers is applied to encode the tongue features and generate the herbal prescription (without considering the dose).
2) Considering the restrictive effect of extremely sparse output labels on the diversity of generated prescriptions, the auxiliary loss of latent therapy topics is proposed to improve the precision and stimulate diversity in prescription construction.
3) We conduct experiments and verify the convolution network for tongue feature encoding and herbal prescription construction. We compare the effects of single/double channel models and the auxiliary topic loss mechanism for accuracy, completeness, and diversity of prescription construction.

The rest of this paper is organized as follows. In Section II, we detail the materials and methods. Sections III and IV illustrate, analyze, and discuss the results of all prescription construction models. We conclude this paper in Section V.

## II. MATERIALS AND METHOD

### A. Collection of Tongue Images and Prescriptions

All of the patients' tongue images are manually taken from the outpatient departments of the cooperating hospital, on the basis of the signed agreements (the number is 9585 persons/times). The shooting tools are digital cameras or smart phones with a more universal image quality.

Handwritten Chinese herbal prescriptions were also taken by volunteers and then entered into the database by professionals. The prescription doses were temporarily masked, and only the names of the herbs were kept.

### B. Architecture of Neural Prescription Construction

*1) Prescription Construction Task Description:* All utilized Chinese herbal medicines were included in a unified dictionary $H = \{h_1, h_2, \ldots, h_n\}$. After we extracted the tongue image features, the task of herbal prescription construction can be seen as an ensemble binary classification task. It needs the tongue image feature vector $f_z$ as an input and outputs the decision vector $P_z = [p_1, p_2, \ldots, p_n]$, whose dimension $n$ is equivalent to the size of the herbal medicine dictionary. The value of element $p_i$ in $P_z$ can only be "1" or "0," which indicates whether the corresponding herb $h_i$ should be included in the prescription construction result.

*2) Convolutional Channels and MLP Encoding Layers:* The convolution network has been successfully applied to the processing and analyses of medical images [20], [21], [36], [37]. As the inputs of tongue image $X_i$ are $(224 \times 224 \times 3)$ matrices with the RGB color channel, the deep convolutional network is used to extract the latent features from the original RGB image matrix, and the multilayer convolutional structure can be defined as a convolutional channel.

The convolutional channel is constructed by several duplicate modules that include convolutional and max-pooling layers. The convolution operation parameterized by the kernel weight matrices $k$ uses the kernel filters to scan the image matrices and reconstruct the feature maps $\mathcal{C}$. The convolution operations in the convolutional channels are abstracted as the following functions with ReLU activation functions:

$$C(X, k) = \text{ReLU}(\text{Conv}(X, k)). \quad (1)$$

With the aim of capturing the most representative features and reducing the computational complexity, max-pooling operations are used to down-sample the feature maps $\mathcal{C}$, as follows:

$$\hat{C}(X, k) = \text{Max}(C(X, k)). \quad (2)$$

Two fully connected layers parameterized by $W_i$ and $W_o$ are used to further encode features as follows:

$$f_z(\hat{C}, \{W_i, W_o\}) = \text{ReLU}(\text{ReLU}(\hat{C}, W_i), W_o). \quad (3)$$

The last layer is also fully connected with weights $W_z$ and the sigmoid activation function. It outputs the probability of

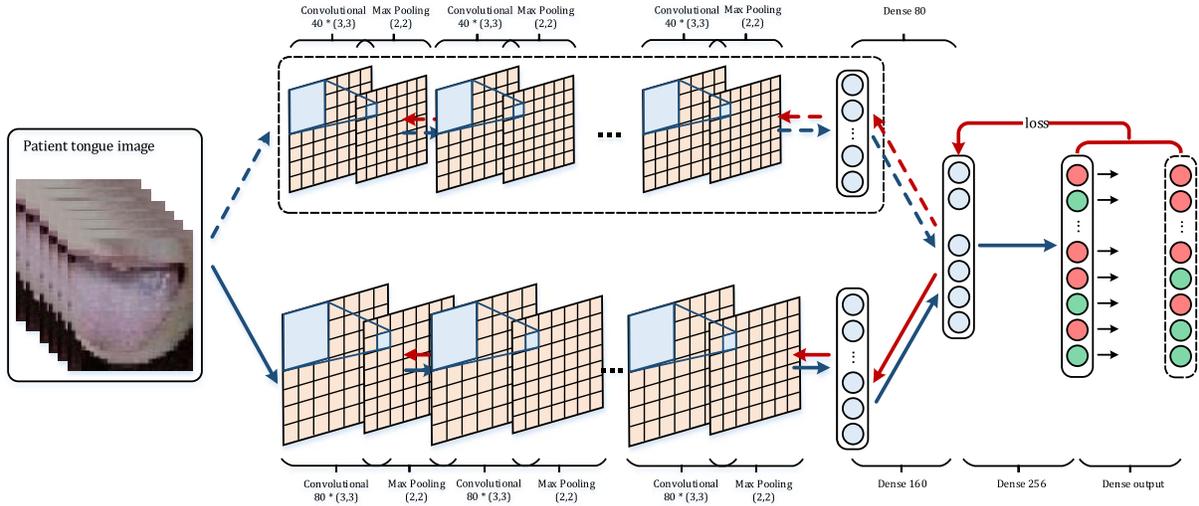

Fig. 2. Architecture of the neural prescription construction model with tongue image inputs.

each herb as follows:

$$P(X, \theta) = \phi(f_z, W_z)$$
$$= [p(h_1|X,\theta), p(h_2|X,\theta), \ldots, p(h_n|X,\theta)] \quad (4)$$

where $\theta = \{k, W_i, W_o, W_z\}$ is the set of all parameters.

Then, the loss function is defined as the mean of the binary cross-entropy for every herb with ground truth $G = [g_1, g_2, \ldots, g_n] g_i \in \{0, 1\}$, and the object minimizes the loss by optimizing all parameters in set $\theta$, as follows:

$$J(\theta) = D_b(P(X, \theta), G)$$
$$= \frac{1}{n}\sum_{i=1}^{n} -g_i \log(p(h_i|X,\theta))$$
$$- (1-g_i)\log(1 - p(h_i|X,\theta)) \quad (5)$$
$$\theta^* = \{k^*, W_i^*, W_o^*, W_z^*\} = \arg\min_\theta D_b(P(X,\theta), G). \quad (6)$$

In many circumstances, Chinese doctors can directly prescribe herbs based on their rich experience, but sometimes they still need to draw support from the diagnosis of the patient's condition to guide the direction of the prescription. Based on this consideration, an auxiliary convolutional channel is added to the prescription construction model. It is used to simulate the diagnosis process and provide more information for the construction of herbal prescriptions.

The structure of the auxiliary convolutional channel is similar to the main convolution channel with less convolutional kernels. It outputs the auxiliary feature maps $\hat{C}_{\text{aux}}$ as follows:

$$\hat{C}_{\text{aux}}(X, k_{\text{aux}}) = \text{Max}(C_{\text{aux}}(X, k_{\text{aux}}))$$
$$= \text{Max}(\text{ReLU}(\text{Conv}_{\text{aux}}(X, k_{\text{aux}}))). \quad (7)$$

Two fully connected layers encode the feature maps from the main and auxiliary convolutional channels and merge them into another fully connected layer. They are parameterized by $W_i$, $W_{\text{aux}}$, and $W_o$, and use the ReLU activation functions as follows:

$$f_i(\hat{C}, W_i) = \text{ReLU}(\hat{C}, W_i) \quad (8)$$

$$f_{\text{aux}}(\hat{C}_{\text{aux}}, W_{\text{aux}}) = \text{ReLU}(\hat{C}_{\text{aux}}, W_{\text{aux}}) \quad (9)$$
$$f_z(\{f_i, f_{\text{aux}}\}, W_o) = \text{ReLU}(f_i \oplus f_{\text{aux}}, W_o) \quad (10)$$

where $\oplus$ is the concatenation operator.

The sigmoid function is still used as the activation function of the output layer, which is the same as (4). The set of parameters is updated as $\theta = \{k, k_{\text{aux}}, W_i, W_{\text{aux}}, W_o, W_z\}$. Then, the objective function is as follows:

$$\theta^* = \{k^*, k_{\text{aux}}^*, W_i^*, W_{\text{aux}}^*, W_o^*, W_z^*\}$$
$$= \arg\min_\theta D_b(P(X,\theta), G). \quad (11)$$

The architecture of the neural prescription construction model is shown in Fig. 2. It consists of one or two convolutional channels and the MLP encoding layers. The outputs of the last layer are sampled from the herb dictionary $H$ through the threshold $t_h$ to obtain the final prescription.

*3) Auxiliary Topic Distribution Ground Truth of Prescription:* The "0," which accounts for the vast majority of output labels and may suppress the confidence of the neural network to produce more diverse and correct herbs. Moreover, besides the real prescriptions ground truth, more auxiliary ground truths are needed to help the network framework model the TCM therapies.

Considering that the prescription does not have obvious context, it can be regarded as a textual model based on a "bag of words." The latent Dirichlet allocation (LDA) model is a classical topic modeling approach based on a bag of words, and it has been successfully applied in the modeling of prescriptions [38].

Using only the training set to model the LDA prescription topics does not affect the primitiveness and objectivity of the test data. Assume that each topic is regarded as a TCM therapy.

The list of TCM therapies is denoted as $K = [k_1, k_2, \ldots, k_m]$, and the Dirichlet prior parameters are $\alpha$ and $\beta$. The LDA model uses Gibbs sampling [42] to estimate the probability that a TCM therapy $k$ is assigned to the $i$th herb $h_i$ in a prescription $P$. This probability in each iteration of Gibbs

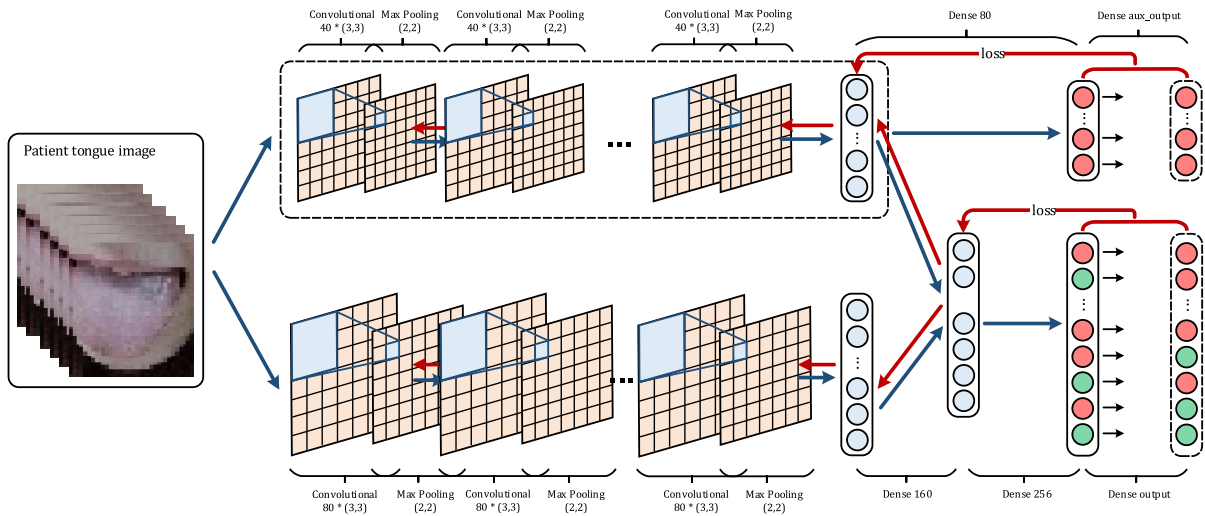

Fig. 3. Architecture of the neural prescription construction 2-output model with tongue image inputs, which is equipped with the auxiliary ground truth of the LDA prescription topics distribution.

sampling is given by

$$p\left(z_{Pi} = k \middle| \vec{z}_{\neg i}\right) \propto \frac{n_{Pk} + \alpha}{\sum_{t \in K} n_{Pt} + |K| \cdot \alpha}$$
$$\times \frac{n_{kh_i} + \beta}{\sum_{j=1}^{|H|} n_{kh_j} + |H| \cdot \beta} \quad (12)$$

where $z_{Pi}$ is the therapy assignment of herb $h_i$ in prescription $P$; $\vec{z}_{\neg i}$ is all herbs' therapy assignments excluding the current one $h_i$; $n_{kh_i}$ is the number of times herb $h_i$ has been assigned to therapy $k$; and $n_{Pk}$ is the number of times herbs in $P$ have been assigned to therapy $k$. Then, the probability distribution of the therapy assignment for prescription $P$ is given by

$$p_{P \to k}(z_P = k|x = P) = \frac{n_{Pk} + \alpha}{\sum_{t \in K} n_{Pt} + |K| \cdot \alpha}. \quad (13)$$

The Kullback–Leibler (KL) divergence [43] can be used to measure the gap between the LDA therapy distributions of generated prescriptions and real prescriptions. Also, the therapy distribution divergence between a real prescription and itself is set as 0 as follows:

$$\bar{D}_{\text{KL}}(P, G)$$
$$= \frac{1}{|K|} \cdot \sum_{k \in K} p_{P \to k}(z_P = k|x = P) \log \frac{p_{P \to k}(z_P = k|x = P)}{p_{G \to k}(z_G = k|x = G)}$$
$$\quad (14)$$
$$\bar{D}_{\text{KL}}(G, G) = 0. \quad (15)$$

The auxiliary convolutional channel [shown as (7) and (9)] outputs the predicted therapy distribution using a fully connected output layer with weight $W_{z\_\text{aux}}$ and the softmax activation function as follows:

$$P_{\text{aux}}(X, \theta_{\text{aux}}) = \text{Softmax}(f_{\text{aux}}, W_{z\_\text{aux}})$$
$$= \begin{bmatrix} p(z = k_1|X, \theta_{\text{aux}}) \\ p(z = k_2|X, \theta_{\text{aux}}) \\ \vdots \\ p(z = k_m|X, \theta_{\text{aux}}) \end{bmatrix}^\top \quad (16)$$

where $f_{\text{aux}}$ is the features encoded by the auxiliary convolutional channel as (7) and (9) and the parameter set for the auxiliary output is denoted as $\theta_{\text{aux}} = \{k_{\text{aux}}, W_{\text{aux}}, W_{z\_\text{aux}}\}$. The loss function of the auxiliary output is given by

$$J(\theta_{\text{aux}}) = \bar{D}_{\text{KL}}(P(X, \theta_{\text{aux}}), G)$$
$$= \frac{1}{|K|} \cdot \sum_{k \in K} p(z = k|X, \theta_{\text{aux}})$$
$$\times \log \frac{p(z = k|X, \theta_{\text{aux}})}{p_{G \to k}(z_G = k|x = G)}. \quad (17)$$

Correspondingly, the loss function of the main output is similar to (4) and parameterized by the main parameter set $\theta_{\text{main}} = \{k, k_{\text{aux}}, W_i, W_{\text{aux}}, W_o, W_z\}$ as follows:

$$J(\theta_{\text{main}}) = D_b(P(X, \theta_{\text{main}}), G). \quad (18)$$

In this way, the LDA topic distribution ground truths of prescriptions are joined to optimize the weights of the auxiliary convolutional channel. This will help the double-output neural network to better model the doctors' therapies, as shown in Fig. 3.

The loss function of the double-output network is integrated by the loss of the auxiliary and main outputs, and the coefficient $\lambda$ is used to regulate the proportion of auxiliary loss. Then, the union set $\theta = \theta_{\text{main}} \bigcup \theta_{\text{aux}}$ in the final objective function is the set of all parameters to be optimized as follows:

$$J(\theta_{\text{main}}, \theta_{\text{aux}}) = D_b(P(X, \theta_{\text{main}}), G) + \lambda \cdot \bar{D}_{\text{KL}}(P(X, \theta_{\text{aux}}), G) \quad (19)$$

$$\theta^* = \theta^*_{\text{main}} \bigcup \theta^*_{\text{aux}}$$
$$= \arg\min_\theta \left( D_b(P(X, \theta_{\text{main}}), G) + \lambda \cdot \bar{D}_{\text{KL}}(P(X, \theta_{\text{aux}}), G) \right). \quad (20)$$

*4) Data Augmentation for Tongue Images:* In the real-world use cases, data collection of patients' tongue images is extremely expensive. Therefore, it is necessary to make the most out of the limited data.

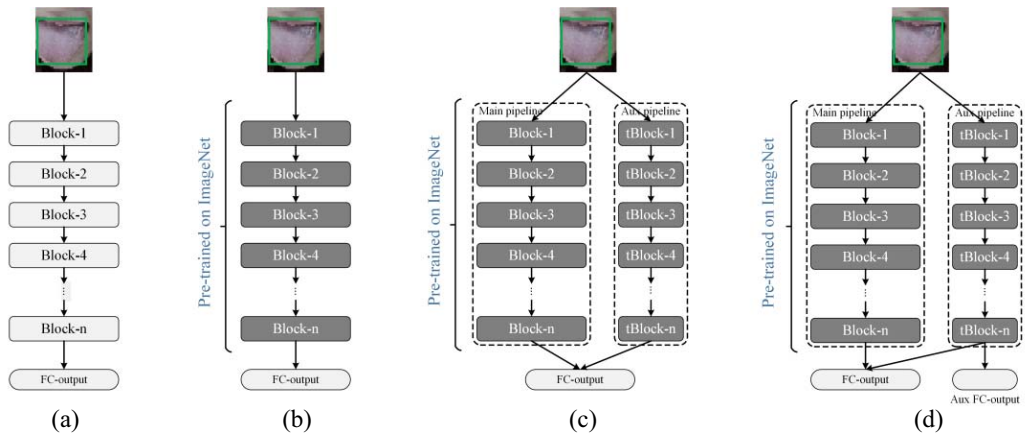

Fig. 4. Deeper prescription construction structures with pretraining on ImageNet. (a) Deep convolutional network with 1 pipeline, which is trained from scratch. (b) Deep convolutional network with 1 pipeline and is pretrained with ImageNet. (c) Deep convolutional network with double pipeline and is pretrained with ImageNet, in which, the "tBlock" refers to the tiny block, it only contains half number of filters each layer than layers in regular block. (d) Deep convolutional network with double pipeline and is pretrained with ImageNet, in which, the auxiliary pipeline also outputs the auxiliary therapy topics.

TABLE I
RESTRICTIVE PARAMETERS FOR IMAGE AUGMENTATION

| Parameters | Explanations | Values |
|---|---|---|
| rotation_range | A value in degrees (0-180) within which pictures are randomly rotated | 25 |
| width_shift_range | Ranges (as a fraction of total width or height) within which to randomly translate pictures vertically or horizontally | 0.05 |
| height_shift_range | | 0.05 |
| shear_range | A value for randomly applying shearing transformations | 0.05 |
| zoom_range | A value for randomly zooming inside pictures | 0.2 |
| horizontal_flip | A flag for randomly flipping half of the images horizontally | True |
| fill_mode | The strategy used for filling in newly created pixels | "nearest" |
| others | | (default) |

The image data generator is used for the data augmentation of tongue images, which can randomly copy and transform the original tongue images. In each iteration, the image generator randomly selects some images from the original images. It then randomly transforms and returns them to the training set. The augmented images keep their original labels. In order to concurrently ensure image quality and data diversity, some restrictive parameters for the image augmentation are set, as shown in Table I.

*5) Weights Transfer With Pretraining on ImageNet:* Moreover, we plan to use the deeper networks as the backbones, to further improve the performance of auto prescription construction. In addition, in order to alleviate the over-fitting caused by nonmassive training data, we pretrain the deeper prescriptions construction models on ImageNet2012 [49] dataset and transfer the weights to training on tongue images.

As shown in Fig. 4, which give the architectures of auto prescription construction models, with deeper backbones, such as VGG16, VGG19 [39], and ResNet50 [41]. We also applied the auxiliary channel design for therapy topics on these deeper networks, implement it with another auxiliary pipeline, which is same structured with the main pipeline and only has half number of filters. All of these settings strictly follow the original design on shallow networks.

So, the main objects of comparison are still the following: 1 pipeline models; 2 pipelines models; and the models with 2 pipelines and auxiliary therapy topics outputs.

In the step of pretraining, all 2 pipelines models are trained as a whole with the shared inputs of ImageNet dataset and without the auxiliary output branch.

## III. RESULTS

### A. Datasets and Preprocessing

The real-world "tongues-prescriptions" experiment dataset that was collected by volunteers contains 9585 tongue images and their corresponding prescriptions. Since the tongue images and prescriptions are very precious, pseudo fivefold cross-validation is used in the experiment. More specifically, 500 samples are randomly selected each time as the test set, the others construct the training set (10% of them are set as the validation set), and the 5 selected test sets do not contain any of the same samples with each other.

The training set can be augmented by the image generator described in Section II-B4. In each iteration, the image generator randomly transforms 64 original images and returns them to the training set. By repeating this operation 200 times, we get 12 800 augmented tongue images.

Over all the prescriptions, there are 566 types of herbs, and the training sets and test sets share one vocabulary for all herbs.

The preprocessing for the tongue images and corresponding prescriptions includes the following two steps.
1) Enable the models to process the directly captured tongue images and only perform slight preprocessing for some low-quality tongue images, including tongue positioning and cutting of nontongue regions.
2) Normalization of the names of herbs. Each herb retains only one unique appellation, and the others are removed.

TABLE II
EXPERIMENT DATASETS ADDITIONAL INFORMATION

| Tongue images info | |
|---|---|
| Size of each training set | 9085 |
| Size of each training set after data augmentation | 21885 |
| RGB image quality | (224× 224×3) |
| Prescriptions info | |
| Maximum number of herbs in a prescription | 28 |
| Minimum number of herbs in a prescription | 2 |
| Average number of herbs in a prescription | 13.89 |
| Average times an herb appears in all prescriptions | 235.25 |
| Proportion of herbs that appear in more than 100 prescriptions (%) | 34.98 |

TABLE III
HYPER-PARAMETERS IN EXPERIMENTAL NEURAL NETWORK MODELS

| Parameters in convolutional channels | |
|---|---|
| Number of kernels of each layer in the main convolutional channel | 80 |
| Number of kernels of each layer in the auxiliary convolutional channel | 40 |
| Kernel size of each layer in the main and auxiliary convolutional channels | (3, 3) |
| Pooling size in the main and auxiliary convolutional channels | (2, 2) |
| Activation for the main and auxiliary convolutional channels | ReLU |
| Batch normalization for the main and auxiliary convolutional channels | True |
| Number of convolution and pooling layer combinations for the main and auxiliary convolutional channels | 3 |
| Parameters in the fully connected encoding layers | |
| Number of neurons in the main fully connected encoding layer | 160 |
| Number of neurons in the auxiliary fully connected encoding layer | 80 |
| Number of neurons in the fully connected merge layer | 256 |
| Dropout rate of the main, auxiliary and merge fully connected layers | 0.5, 0.5, 0.6 |
| Activation for the main, auxiliary and merge fully connected layers | ReLU |
| Activation for the main output layer | Sigmoid |
| Activation for the auxiliary output layer | Softmax |

Additional information about the datasets is listed in Table II.

### B. Experimental Setup

*1) Experimental Workflow:* To the best of our knowledge, there has not been any other work that strictly explores the same task as this paper. The baseline method is set as directly using the random forest multilabel image classifier on herbs.

The experiment will also examine the effectiveness of the prescription construction ability of bi-convolutional channels, auxiliary therapy topics and image augmentation. Brief introduction of the comparison methods are as follows.

*RF-Baseline:* Implement the multilabel classification on herbs and tongue images using the multilabel supported random forest classifier [44].

*1-CNNs + MLP:* The neural prescription construction model contains only one convolutional channel and fully connected encoding layers. Also, the sigmoid function is used in the output layer.

*2-CNNs + MLP:* It is like the previous model, and the only difference is that it contains two convolutional channels.

*2-CNNs + MLP + AUX_LDA:* Based on the model "2-CNNs + MLP," the therapies are modeled by introducing the auxiliary ground truths of the prescription topics, and the auxiliary convolution channel is optimized.

*1-CNNs + MLP With Data Augmentation:* It is like the model "1-CNNs + MLP," but it uses the augmented dataset.

*2-CNNs + MLP With Data Augmentation:* It is like the model 2-CNNs + MLP, but it uses the augmented dataset.

*2-CNNs + MLP + AUX_LDA With Data Augmentation:* It is like the model "2-CNNs + MLP + AUX_LDA," but it uses the augmented dataset.

*2) Parameter Setting of Neural Networks:* In order to take into account the performance of the model and the consumption of computing resources, the parameters of the neural prescription construction models are set by the experience of many experimental iterations. The main parameters are listed in Table III.

*3) Evaluation Metrics:* To measure the similarity between the generated prescription and the real prescription, the metrics precision similarity $p\_sim$, recall similarity $r\_sim$, and intersection-over-union rate $IoU\_sim$ are set as follows:

$$p\_\text{sim} = \frac{1}{N_t}\sum_{i=1}^{N_t}\frac{nc_i}{np_i}, r\_\text{sim} = \frac{1}{N_t}\sum_{i=1}^{N_t}\frac{nc_i}{ng_i}$$

$$\text{IoU\_sim} = \frac{1}{N_t}\sum_{i=1}^{N_t}\frac{nc_i}{np_i + ng_i - nc_i} \quad (21)$$

where $nc_i$ is the number of identical herbs in the generated prescription and the real prescription of the $i$th sample; $np_i$ and $ng_i$ are the respective numbers of herbs in the generated prescription and the real prescription of $i$th sample; and $N_t$ is the number of test samples. In addition, $IoU\_sim$ cannot be directly deduced from $p\_sim$ and $r\_sim$ at the macro level.

Sometimes, the above three metrics cannot perfectly evaluate the performance of prescription construction. Some additional metrics are taken into account, which are used to indicate the completeness of the prescription and the exactitude of the therapies. They are as follows.

*nb_p:* The average number of herbs that appear in the generated prescriptions.

*nb_c:* The average number of correct herbs that appear in the generated prescriptions.

*nb_d:* The absolute value of the average quantity difference of herbs that appear in the generated prescriptions and the real prescriptions.

*kl_t:* The KL divergence of topic distributions from the generated prescription to the real prescription.

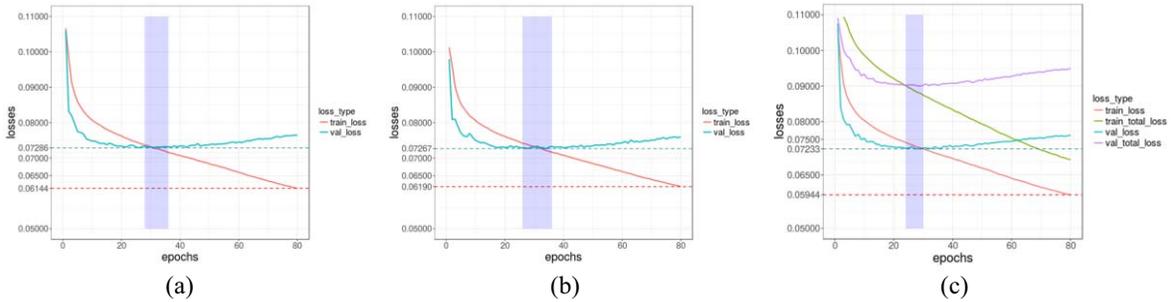

Fig. 5. Validation losses for each model with respect to the training epochs without data augmentation. (a) 1-CNNs + MLP; (b) 2-CNNs + MLP; and (c) 2-CNNs + MLP + AUX_LDA. The red line represents the main loss, and the green line represents the total loss. The red shadow indicates the confidence field of the validation loss, and the blue shadow indicates the range of training epochs that reach the least validation loss.

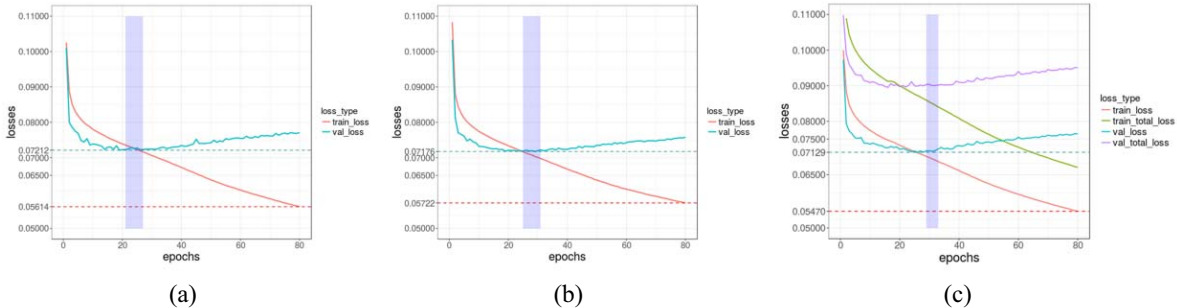

Fig. 6. Validation losses for each model with respect to the training epochs with data augmentation. (a) 1-CNNs + MLP; (b) 2-CNNs + MLP; and (c) 2-CNNs + MLP + AUX_LDA. The red line represents the main loss, and the green line represents the total loss. The red shadow indicates the confidence field of the validation loss, and the blue shadow indicates the range of training epochs that reach the least validation loss.

## C. Model Evaluation

*1) Training Process and Therapy Topics Distributions:* Figs. 5 and 6 show the trend of validation losses with regard to the training process of the neural prescription construction models. From Fig. 5, we see that the variations and the min-values of the validation losses for the different models are very close. However, the training process with data augmentation can obtain relatively lower validation loss for each model, and the 2-CNNs + MLP + AUX_LDA model obtains the lowest validation loss regardless of whether data augmentation is used. In particular, the optimal performance of the negative log-likelihood for the 2-CNNs + MLP + AUX_LDA model without data augmentation is close to the 1-CNNs + MLP model after using data augmentation.

The green lines in Figs. 5(c) and 6(c) show a fairly similar trend to the validation loss, which indicates the integration of the negative log-likelihood of prescriptions and the auxiliary loss of therapy topics. Therefore, the learning objective of the auxiliary output is consistent with the generation of the prescription, which can promote the optimization of the prescription construction models.

The blue shadows in Figs. 5 and 6 show that the training of the model will converge in the 20th to 30th round. The wider shadow reveals the greater fluctuation of the convergence time in the fivefold experiments, and the convergence time of the training with data augmentation is generally more stable. Moreover, in training processes without data augmentation, the 2-CNNs + MLP + AUX_LDA model is most stable. On another note, the confidence field of each model reveals that the training of the 2-CNNs + MLP + AUX_LDA model has smaller undulations in all cases.

Fig. 11, in the supplementary material, shows the complete topic distributions that are (all compressed to 2-dim by online dictionary learning function [45] in the fivefold experiments for all neural prescription construction models. For a more convenient observation, Fig. 7 illustrates the comparative examples of the therapy topic distributions from the 1-CNNs + MLP without data augmentation and 2-CNNs + MLP + AUX_LDA with data augmentation models. The figure clearly exhibits the differences in therapy distributions between an ordinary result and an optimal result.

The therapy topic distribution of the real samples shows some concentration. This means that there are some frequently used herbs in the real prescriptions, and the prescriptions composed of some seldom-used herbs are sparsely distributed in the marginal area. For the 1-CNNs + MLP without data augmentation model, the topic distribution of its generated samples partially fit the real samples, but they are more concentrated, which reveals that these generated prescriptions are more frequent for the use of common herbs. Moreover, a part of these results is gathered in the marginal regions that are not fit for the real samples, and most of the outlier real samples are not fitted. In contrast, for the 2-CNNs + MLP + AUX_LDA with data augmentation model, although the topic distribution of the generated samples is still a little more concentrated than the real samples, it appears quite homogeneous and well fitting, regardless of whether it is in the concentrated area of the real samples or the slightly sparse area. Even in the vicinity of the outlier real samples, there are still some generated samples that try to fit them. Therefore, the diversity of the generated samples can be reflected, which is closer to the real prescription samples.

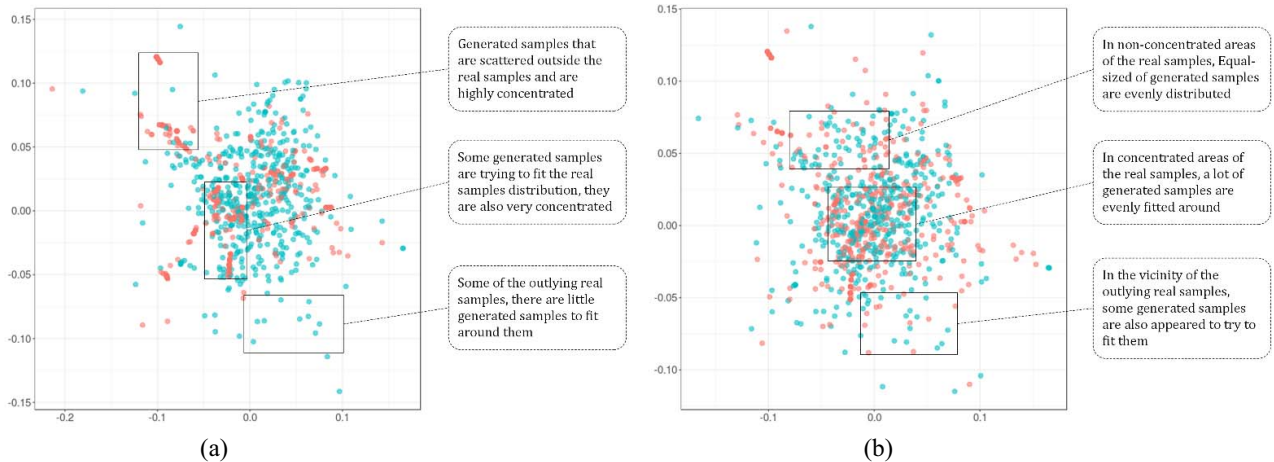

Fig. 7. Comparative examples of the topic distributions (all compressed to 2-dim) of various models' prescription construction results. (a) 1-CNNs + MLP without data augmentation and (b) 2-CNNs + MLP + AUX_LDA with data augmentation. The red and green dots represent the generated and real samples, respectively.

*2) Comparison Results:* As shown in Table IV, the performances of all neural prescription construction models are better than the RF-baseline, which shows the result of the collapse of the metric IoU_sim. When data augmentation is not used, the performance of the 1-CNNs + MLP, 2-CNNs + MLP, and 2-CNNs + MLP + AUX_LDA models is increasing. The improvement of the 2-CNNs + MLP + AUX_LDA model on the *p*_sim and *r*_sim metrics is more obvious, and its performance is even better than the 1-CNNs + MLP model with data augmentation. When data augmentation is used, the performances of various models is generally enhanced due to the increased training samples. Of note, the AUX_LDA mechanism has a significant effect on the model's performance improvement and the 2-CNNs + MLP + AUX_LDA with data augmentation model achieves the best performance on the main metrics.

In some conditions, part of the main metrics cannot well reflect the model's prescription construction ability well. For example, too few and coincidental correct herbs in the generated prescription will lead to the false high *p*_sim, and too many herbs in a generated prescription may lead to the false high *r*_sim.

As the supplement to Table IV, Table V shows that the average size *nb*_p of the generated prescriptions is less than the real prescriptions. The majority of the "0"s in the output labels inhibit the confidence of the prescription construction model, and they prefer to prescribe less herbs rather than take the risk of prescribing the wrong herb. By contrast, the size of the prescription produced by the 2-CNNs + MLP + AUX_LDA with the data augmentation model is the closest to the Real. Furthermore, the AUX_LDA mechanism that encourages the models to prescribe more herbs can also effectively improve the number of correct herbs *nb*_c. As described in Section III-C1 with the LDA topic distributions of the prescriptions, the 2-CNNs + MLP + AUX_LDA model can also acquire the better performance on metric *kl*_t regardless of the use of data augmentation.

Tables VIII and IX more vividly illustrate the prescription construction results of various models in the experiment. It can be seen that the 2-CNNs + MLP model can more accurately prescribe herbs than the 1-CNNs + MLP model. Furthermore, the 2-CNNs + MLP + AUX_LDA model can accurately prescribe less-common herbs, even if a few wrong herbs are occasionally prescribed, such as "Sculellaria barbata" and "Semen zizyphi spinosae" (in the sample of Table VIII and the sample of Table IX). Interestingly, some of the wrong herbs prescribed by the 2-CNNs + MLP + AUX_LDA model have similar benefits as some missing correct herbs, like "Herba Hedyotis" and "Sculellaria barbata" (in the sample of Table VIII).

*3) Artificial Evaluation:* We also introduce the artificial evaluation for auto-generated prescriptions, to check the Chinese pharmacological logic of them. We invited more than 10 doctors from Department of TCM in Guangdong General Hospital, to conduct artificial evaluation for auto-generated prescriptions of 500 samples. The main content of the artificial evaluation is checking the rationality of the compatibility logic of the prescription, it mainly includes two factors: 1) whether the herbs in same prescription constitute some common-pairs and 2) whether there is any conflict in same prescription. As shown in Table VI, the green boxes mark the common-pairs of herbs, which indicate that these herbs are commonly used together to play a mutually beneficial role, and the red frames mark the conflict-pairs, which mean that these herbs are not suitable for use together.

Although the logical evaluation of the prescription is solely based on the subjective judgement of the doctors, it is basically based on the following rules:

$$S_l(p_i) = S_{\text{pos}}(p_i) + S_{\text{neg}}(p_i) \tag{22}$$

where $S_l(\cdot)$ refers to the total score of logic evaluation of a prescription; $S_{\text{pos}}(\cdot)$ and $S_{\text{neg}}(\cdot)$ are the scores of common herb pair and avoidance for taboo conflict, the results of artificial evaluation are listed in Table VII. Although the ideal artificial evaluation score needs as large as possible, almost all the scores of artificial evaluation are within this range: (0, 25].

TABLE IV
COMPARISON RESULTS FOR THE PRESCRIPTION CONSTRUCTIONS OF VARIOUS MODELS WITH THE MAIN METRICS.
HIGHER VALUES OF THESE METRICS INDICATE A BETTER PERFORMANCE OF PRESCRIPTION CONSTRUCTION

|  | p_sim (%) | r_sim (%) | IoU_sim (%) |
| --- | --- | --- | --- |
| RF-baseline | 39.83 ± 1.17 | 28.96 ± 0.96 | 21.89 ± 1.27 |
| **1-CNNs + MLP** | 43.26 ± 1.06 | 32.89 ± 1.18 | 28.14 ± 1.06 |
| **2-CNNs + MLP** | 44.39 ± 1.35 | 33.55 ± 1.30 | 28.61 ± 1.35 |
| **2-CNNs + MLP + AUX_LDA** | 45.28 ± 0.64 | 35.62 ± 1.36 | 29.72 ± 0.64 |
| **1-CNNs + MLP** with data augmentation | 45.27 ± 0.63 | 34.55 ± 1.42 | 28.97 ± 0.63 |
| **2-CNNs + MLP** with data augmentation | 45.52 ± 0.97 | 35.84 ± 1.13 | 30.62 ± 0.97 |
| **2-CNNs + MLP + AUX_LDA** with data augmentation | **46.77 ± 1.03** | **37.17 ± 0.88** | **32.23 ± 1.03** |

TABLE V
COMPARISON RESULTS FOR THE PRESCRIPTION CONSTRUCTIONS OF VARIOUS MODELS WITH ADDITIONAL METRICS. THE CLOSER
TO THE REAL VALUE OF **NB_P**, THE HIGHER IS THE VALUE OF **NB_C**, AND LOWER VALUES OF **NB_D** AND **KL_T**
INDICATE BETTER PRESCRIPTION CONSTRUCTION PERFORMANCE

|  | nb_p | nb_c | nb_d | kl_t |
| --- | --- | --- | --- | --- |
| *Real* | 13.85 | - | - | - |
| RF-baseline | 7.18 | 2.76 | 8.67 | 3.41 |
| **1-CNNs + MLP** | 9.53 | 4.28 | 4.94 | 2.79 |
| **2-CNNs + MLP** | 9.64 | 4.43 | 4.81 | 2.74 |
| **2-CNNs + MLP + AUX_LDA** | 10.61 | 4.99 | 4.04 | 2.59 |
| **1-CNNs + MLP** with data augmentation | 10.30 | 4.57 | 3.95 | 2.61 |
| **2-CNNs + MLP** with data augmentation | 11.14 | 5.12 | 3.11 | 2.54 |
| **2-CNNs + MLP + AUX_LDA** with data augmentation | 11.55 | 5.53 | 2.68 | 2.39 |

TABLE VI
EXAMPLES OF LOGIC PAIRING IN HERBAL PRESCRIPTIONS. THE RED
MARKS REFER TO THE CONFLICT-PAIRS AND THE GREEN MARK
INDICATE THE COMMON COOPERATIVE PAIRS

| Generated prescription |
| --- |
| 柴胡 白芍 香附 枳壳 党参 白术 海螵蛸 姜半夏 炙甘草 紫苏叶 (Bupleurum, Radix paeoniae alba, Cyperus Rotundus, Fructus aurantii, Codonopsis pilosula, Rhizome of lagehead atractylodes, Cuttle-bone, Pinellia Rhizoma Prepared by Ginger, Caulis perillae acutae, Folia perillae acutae) |
| 桂枝 甘草 党参 太子参 黄芪 附子 珍珠母 (Cassia twig, Liquorice, Codonopsis pilosula, Radix Pseudostellariae, Astragalus membranaceus, Monkshood, Mother of pearl) |

Detailed steps of artificial evaluation and counting method of $S_{\text{pos}}(\cdot)$ and $S_{\text{neg}}(\cdot)$ are as follows.
1) The chief expert formulates the basic principles of common-pairs and conflict which are discussed by all experts.
2) The doctors evaluate the prescriptions according to their medical experience: if a beneficial herb-pair appears, they add 1 to $S_{\text{pos}}(\cdot)$, when a taboo herb-pair appears, 1 is subtracted from $S_{\text{neg}}(\cdot)$, and on the contrary, if a taboo pair is successfully avoided, 1 is added to $S_{\text{neg}}(\cdot)$.
3) The doctors exchange the prescriptions for evaluation, and ensure that each prescription has been evaluated by no less than three experts. In case of inconsistencies in the evaluation, experts immediately discuss and give comprehensive results.

As listed in Table VII, the artificial evaluation scores of the generated prescriptions are still less than that of the corresponding real prescriptions, the main difference is that the generated prescription seldom contains the common herb pair. Then, we can also notice that the mechanism of AUX_LDA can alleviate this problem quite well, the model of 2-CNNs + MLP + AUX_LDA can generate more common-pairs of herbs while avoiding taboo conflicts, thus, the models of 2-CNNs + MLP + AUX_LDA have a relatively higher artificial evaluation scores than other models. Fig. 8 also shows the artificial evaluation scores (herbs logic check scores) of all compared models, for 500 samples. Further, we use linear discriminant analysis (Linear-DA) to reduce the dimension of artificial evaluation score of different models, then metric the distance of each tested model from label prescription. Actually, we want the artificial evaluation score of generated prescriptions can be as high as possible. However, due to the score of the generated prescriptions are commonly lower than that of the real prescriptions, the smaller distance from the real prescriptions indicates the better result. As listed in last column of Table VII, the models using the mechanism of AUX_LDA get the smallest distance from real prescription.

Moreover, we also analyze the score distributions of artificial evaluation, as shown in Fig. 9. The artificial evaluation score distributions is roughly obeying the Gaussian distribution. Where the blue part indicates the label prescriptions from veteran doctors and the orange part refers to the generated prescription from proposed models, axis *X* is the line of score and axis *Y* means the probability on corresponding score. We can notice that the distribution of the label prescription score is usually more than that of the generated prescription score in higher score area. Also, we can see the score distribution of the generated prescription is most similar with that of the label prescription when using the mechanism of AUX_LDA.

TABLE VII
COMPARISON RESULTS FOR THE PRESCRIPTION CONSTRUCTIONS OF VARIOUS MODELS WITH THE MAIN METRICS.
HIGHER VALUES OF THESE METRICS INDICATE A BETTER PERFORMANCE OF PRESCRIPTION CONSTRUCTION

|  | Herb pair | Taboo avoidance | Total score | Linear-DA distance |
|---|---|---|---|---|
| **Label** | 0.792 | 5.866 | 6.658 | — |
| **1-CNNs + MLP** | 0.060 | 3.086 | 3.146 | 4.248 |
| **2-CNNs + MLP** | 0.052 | 3.296 | 3.348 | 3.358 |
| **2-CNNs + MLP + AUX_LDA** | 0.616 | 4.770 | 5.386 | 1.831 |
| **1-CNNs + MLP** with data augmentation | 0.064 | 3.378 | 3.442 | 3.157 |
| **2-CNNs + MLP** with data augmentation | 0.076 | 3.640 | 3.716 | 3.664 |
| **2-CNNs + MLP + AUX_LDA** with data augmentation | 0.558 | 4.528 | 5.086 | 1.783 |

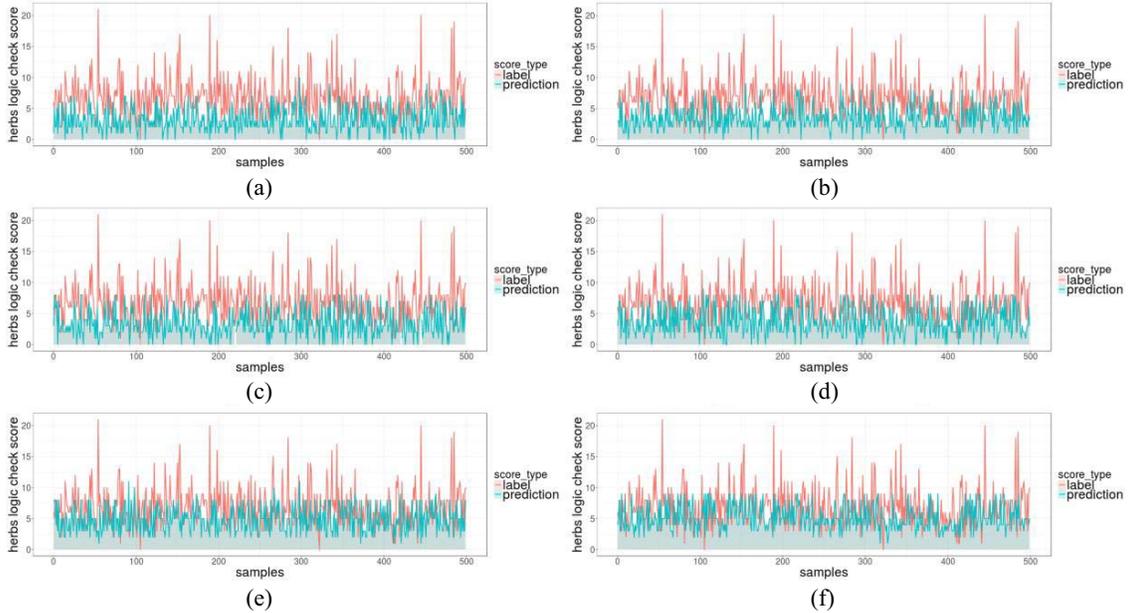

Fig. 8. Comparison results of artificial evaluation score of different models. All the prescriptions are from the same random sampled 500 cases, and the red line refer to the scores of label predictions and the blue lines indicate the scores of generated predictions. (a) 1-CNNs + MLP. (b) 1-CNNs + MLP with data augmentation. (c) 2-CNNs + MLP. (d) 2-CNNs + MLP with data augmentation. (e) 2-CNNs + MLP + AUX_LDA. (f) 2-CNNs + MLP + AUX_LDA with data augmentation.

*4) Results of Deeper Models With Weights Transfer:* Table X lists the results of deeper prescription construction models with pretraining on ImageNet2012 [49], the upper half part indicates the results without data augmentation, and the lower part indicates the results with data augmentation on tongue image datasets. From this table, we can notice that the transferred weights from ImageNet do provide impressive performance improvements, for deeper models in prescription construction, the overfitting phenomenon has also been alleviated to some extent. In addition, with pretrained weights, the deep models can converge faster (within nearly 30 epochs). Last but not least, the design of auxiliary therapy topics framework can further improve the performance of prescription generation, than typical double pipeline deeper models, which means the proposed auxiliary therapy topics module is also effective, on deeper backbones.

## IV. Discussion

Tongue images provide important bodily information and can serve as an important basis for clinical diagnosis and treatment. In this paper, neural network models are used to model the relationship between tongue images and Chinese herbal prescriptions. Then, we automatically construct the herbal prescriptions. After that we verified its validity through experiments. The performance of the proposed method is outstanding in a very small number of similar studies [48]. However, the tongue image is not the only basis for a doctor to make a diagnosis and treatment. It is reasonable to believe that integrating more body sign information can further improve the performance of automatic prescription construction, and this paper provides a reference for tongue images.

The model of the double convolution channel can improve the precision of prescription construction, and the auxiliary therapy topic output mechanism can further improve the precision and diversity of the generated prescription. In addition, we fuse the features of double channels before output, in order to retain the original information of the two channels as more as possible and avoid unnecessary loss of features, we choose the strategy of "concatenation" to merge the data flows.

We further validate the effectiveness of the proposed design of auxiliary therapy topic, on deeper backbones, like VGG [39] and residual network [41].

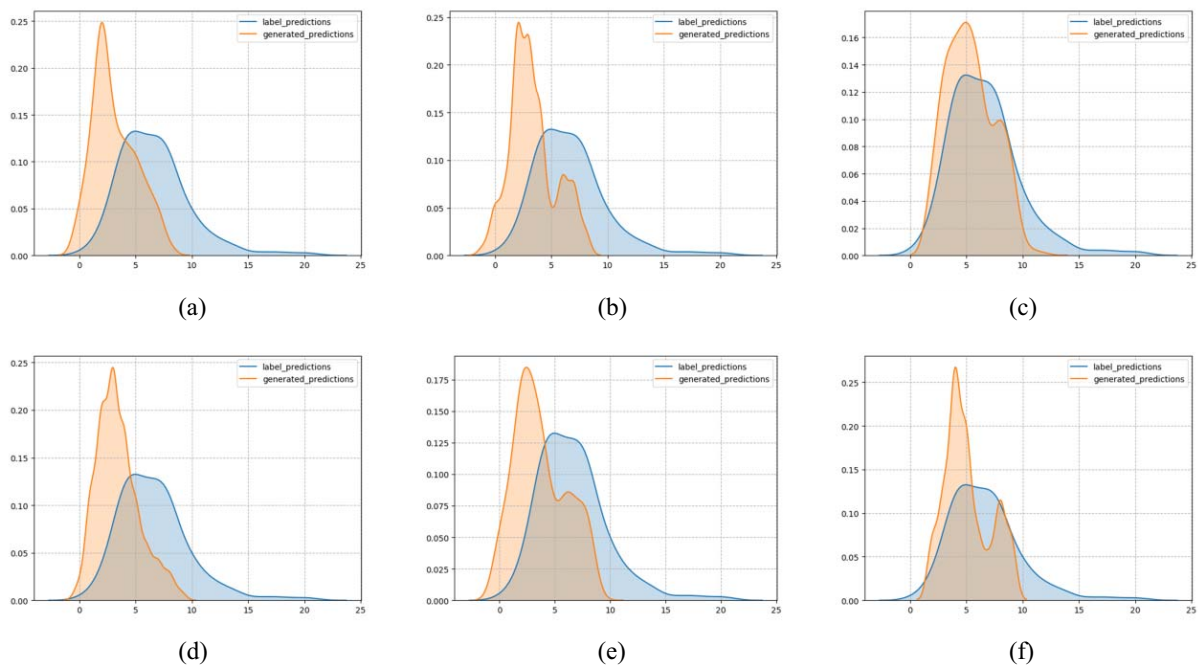

Fig. 9. Artificial evaluation score distribution compare of generated prescriptions and label prescriptions. (a) 1-CNNs + MLP. (b) 2-CNNs + MLP. (c) 2-CNNs + MLP + AUX_LDA. (d) 1-CNNs + MLP with data augmentation. (e) 2-CNNs + MLP with data augmentation. (f) 2-CNNs + MLP + AUX_LDA with data augmentation.

TABLE VIII
PRESCRIPTION CONSTRUCTION EXAMPLES OF VARIOUS MODELS WITHOUT DATA AUGMENTATION. THE GREEN WORDS REPRESENT THE CORRECT HERBS (ALSO APPEAR IN THE CORRESPONDING REAL PRESCRIPTION) IN GENERATED PRESCRIPTION, WHILE THE RED WORDS AND BLACK WORDS IN BOXES REPRESENT THE WRONG HERBS (DO NOT APPEAR IN THE CORRESPONDING REAL PRESCRIPTION) AND THE MISSING HERBS (APPEAR IN THE CORRESPONDING REAL PRESCRIPTION BUT NOT APPEAR IN THE GENERATED PRESCRIPTION)

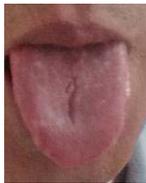

The auxiliary therapy topic output mechanism is sensitive to a few hyper-parameters, and we can determine the optimal selection of these parameters by several enumerated experiments.

Although the experiment in this paper used a dataset of Chinese patients' tongue images and Chinese herbal prescriptions, the proposed model can be easily extended to other datasets and application scenarios.

For two reasons, we do not take the dose into consideration in this paper.

1) The current study is very basic, there is still much room for improving the performance of herbs recommendation.
2) The application set up of this paper is to provide herbal tips for the rookie doctors, so, using or not using of one herb seems more important than its dosage. Although we have not considered the dose for the time being, we still take it into consideration in our future research plan.

TABLE IX
PRESCRIPTION CONSTRUCTION EXAMPLES OF VARIOUS MODELS WITH DATA AUGMENTATION. THE GREEN WORDS REPRESENT THE CORRECT HERBS (ALSO APPEAR IN THE CORRESPONDING REAL PRESCRIPTION) IN GENERATED PRESCRIPTION, WHILE THE RED WORDS AND BLACK WORDS IN BOXES REPRESENT THE WRONG HERBS (DO NOT APPEAR IN THE CORRESPONDING REAL PRESCRIPTION) AND THE MISSING HERBS (APPEAR IN THE CORRESPONDING REAL PRESCRIPTION BUT NOT APPEAR IN THE GENERATED PRESCRIPTION)

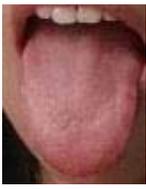

| | | |
|---|---|---|
| | Real | 甘草 茯苓 木香 白术 黄连 山药 天麻 鳖甲 粉葛 蜈蚣 白花蛇舌 天山雪莲 半枝莲 红豆杉 (Liquorice, Poria cocos, Common vladimiria root, Rhizome of lagehead atractylodes, Goldthread, Chinese Yam, Gastrodia tuber, Turtle shell, Kudzu, Centipede, Herba Hedyotis, Saussurea involucrate, Scullellaria barbata, Chinese yew) |
| | 1-CNNs + MLP with data augmentation | 甘草 茯苓 党参 木香 白术 生地黄 黄连 山药 天麻 鳖甲 粉葛 蜈蚣 白花蛇舌 天山雪莲 半枝莲 红豆杉 (Liquorice, Poria cocos, Codonopsis pilosula, Common vladimiria root, Rhizome of lagehead atractylodes, Rehmannia glutinosa, Goldthread, Chinese Yam, Gastrodia tuber, Turtle shell, Kudzu, Centipede, Herba Hedyotis, Saussurea involucrate, Scullellaria barbata, Chinese yew) |
| | 2-CNNs + MLP with data augmentation | 甘草 茯苓 党参 木香 白术 生地黄 黄连 山药 天麻 鳖甲 粉葛 蜈蚣 白花蛇舌 天山雪莲 半枝莲 红豆杉 (Liquorice, Poria cocos, Codonopsis pilosula, Common vladimiria root Rhizome of lagehead atractylodes, Rehmannia glutinosa, Goldthread, Chinese Yam, Gastrodia tuber, Turtle shell, Kudzu, Centipede, Herba Hedyotis, Saussurea involucrate, Scullellaria barbata, Chinese yew) |
| | 2-CNNs + MLP + AUX_LDA with data augmentation | 甘草 茯苓 党参 木香 白术 酸枣仁 生地黄 黄连 山药 天麻 鳖甲 粉葛 蜈蚣 白花蛇舌 天山雪莲 半枝莲 红豆杉 (Liquorice, Poria cocos, Codonopsis pilosula, Common vladimiria root Rhizome of lagehead atractylodes, Semen zizyphi spinosae, Rehmannia glutinosa, Goldthread, Chinese Yam, Gastrodia tuber, Turtle shell, Turtle shell, Kudzu, Herba Hedyotis, Saussurea involucrate, Scullellaria barbata, Chinese yew) |

TABLE X
COMPARISON PRESCRIPTION CONSTRUCTIONS RESULTS FOR THE PROPOSED MODELS USING DEEPER BACKBONES WITH THE MAIN METRICS. HIGHER VALUES OF THESE METRICS INDICATE A BETTER PERFORMANCE OF PRESCRIPTION CONSTRUCTION

| | p_sim (%) | r_sim (%) | IoU_sim (%) |
|---|---|---|---|
| **1-Pipeline VGG16** | 44.28 ± 1.08 | 32.97 ± 1.13 | 28.95 ± 0.94 |
| **2-Pipelines VGG16** | 44.78 ± 1.22 | 33.52 ± 1.04 | 29.01 ± 1.05 |
| **2-Pipelines VGG16 + AUX_LDA** | 46.02 ± 0.97 | 35.99 ± 1.01 | 30.56 ± 0.86 |
| **1-Pipeline VGG19** | 44.30 ± 0.95 | 33.60 ± 1.07 | 29.57 ± 0.93 |
| **2-Pipelines VGG19** | 45.38 ± 1.01 | 33.45 ± 1.02 | 29.76 ± 0.78 |
| **2-Pipelines VGG19 + AUX_LDA** | 45.96 ± 0.80 | **36.16 ± 0.65** | 30.74 ± 0.86 |
| **1-Pipeline ResNet50** | 44.73 ± 1.02 | 33.26 ± 0.82 | 29.81 ± 0.88 |
| **2-Pipelines ResNet50** | 45.30 ± 0.80 | 34.92 ± 0.96 | 30.41 ± 1.19 |
| **2-Pipelines ResNet50+ AUX_LDA** | **46.17 ± 0.81** | 36.13 ± 1.03 | **30.82 ± 1.06** |
| **1-Pipeline VGG16** with data augmentation | 45.98 ± 1.17 | 35.01 ± 0.93 | 30.40 ± 0.79 |
| **2-Pipelines VGG16** with data augmentation | 46.09 ± 1.07 | 35.31 ± 1.06 | 30.78 ± 1.05 |
| **2-Pipelines VGG16 + AUX_LDA** with data augmentation | 46.89 ± 0.88 | 36.30 ± 1.02 | 32.81 ± 0.90 |
| **1-Pipeline VGG19** with data augmentation | 45.36 ± 0.98 | 34.95 ± 0.97 | 31.13 ± 0.77 |
| **2-Pipelines VGG19** with data augmentation | 46.69 ± 0.59 | 36.32 ± 1.03 | 31.81 ± 0.90 |
| **2-Pipelines VGG19 + AUX_LDA** with data augmentation | 47.69 ± 0.88 | 37.59 ± 1.01 | 33.08 ± 0.80 |
| **1-Pipeline ResNet50** with data augmentation | 46.06 ± 0.89 | 35.07 ± 0.63 | 31.64 ± 0.67 |
| **2-Pipelines ResNet50** with data augmentation | 46.94 ± 0.73 | 36.28 ± 1.06 | 32.11 ± 0.91 |
| **2-Pipelines ResNet50+ AUX_LDA** with data augmentation | **48.09 ± 0.77** | **38.02 ± 1.02** | **33.44 ± 0.85** |

Last but not least, the deep learning algorithm, including the proposed models, can be expediently run on a distributed platform [46].

## V. CONCLUSION

This paper attempts to apply neural network models to fulfill the automatic construction of Chinese herbal prescriptions. To improve the accuracy and diversity of prescription construction, the double convolutional channel frame is applied and the auxiliary therapy topic output mechanism is proposed. Then, the abilities of the proposed models are verified in experiments on the automatic construction of Chinese herbal prescriptions.

In future work, we plan to do the following.

1) Collect more data (not only for tongue images) and then verify and improve the performance of the proposed models on larger and richer datasets.
2) Absorb more medicine transcendental knowledge or ontology [35], [47] to guide the neural prescription construction models.
3) Invite experienced Chinese doctors to provide a more professional artificial score for the generated prescription, which would introduce the online reinforcement learning mechanism into the basic system.

ACKNOWLEDGMENT

The authors would like to thank the anonymous reviewers for their constructive advice. They are also grateful for the hard works of the data collection volunteers.

This article has been accepted for inclusion in a future issue of this journal. Content is final as presented, with the exception of pagination.

HU et al.: AUTOMATIC CONSTRUCTION OF CHINESE HERBAL PRESCRIPTIONS FROM TONGUE IMAGES 13## REFERENCES

[1] J. Yang, Y. Hong, and S. Ma, "Impact of the new health care reform on hospital expenditure in China: A case study from a pilot city," *China Econ. Rev.*, vol. 39, pp. 1–14, Jul. 2016.

[2] C. C. Yang and P. Veltri, "Intelligent healthcare informatics in big data era," *Artif. Intell. Med.*, vol. 65, no. 2, pp. 75–77, 2015.

[3] P. Chen et al., "Telehealth attitudes and use among medical professionals, medical students and patients in China: A cross-sectional survey," *Int. J. Med. Informat.*, vol. 108, pp. 13–21, Dec. 2017.

[4] M. Zięba, "Service-oriented medical system for supporting decisions with missing and imbalanced data," *IEEE J. Biomed. Health Informat.*, vol. 18, no. 5, pp. 1533–1540, Sep. 2014.

[5] R. K. Lomotey et al., "Mobile medical data synchronization on cloud-powered middleware platform," *IEEE Trans. Services Comput.*, vol. 9, no. 5, pp. 757–770, Sep./Oct. 2016.

[6] C. Dainton and C. H. Chu, "A review of electronic medical record keeping on mobile medical service trips in austere settings," *Int. J. Med. Informat.*, vol. 98, pp. 33–40, Feb. 2017.

[7] J. C. Tilburt and T. J. Kaptchuk, "Herbal medicine research and global health: An ethical analysis," *Bull. World Health Org.*, vol. 86, no. 8, pp. 594–599, 2008.

[8] Q. Li, Y. Wang, H. Liu, Z. Sun, and Z. Liu, "Tongue fissure extraction and classification using hyperspectral imaging technology," *Appl. Opt.*, vol. 49, no. 11, pp. 2006–2013, 2010.

[9] Q. Li, Y. Wang, H. Liu, and Z. Sun, "AOTF based hyperspectral tongue imaging system and its applications in computer-aided tongue disease diagnosis," *Biomed. Eng. Informat.*, pp. 1424–1427, 2010.

[10] L. Zhuo, J. Zhang, P. Dong, Y. Zhao, and B. Peng, "An SA–GA–BP neural network-based color correction algorithm for TCM tongue images," *Neurocomputing*, vol. 134, pp. 111–116, Jun. 2014.

[11] L. Zhuo et al., "A K-PLSR-based color correction method for TCM tongue images under different illumination conditions," *Neurocomputing*, vol. 174, pp. 815–821, Jan. 2016.

[12] B. Pang, D. Zhang, N. Li, and K. Wang, "Computerized tongue diagnosis based on Bayesian networks," *IEEE Trans. Biomed. Eng.*, vol. 51, no. 10, pp. 1803–1810, Oct. 2004.

[13] X. Wang, B. Zhang, Z. Yang, H. Wang, and D. Zhang, "Statistical analysis of tongue images for feature extraction and diagnostics," *IEEE Trans. Image Process.*, vol. 22, no. 12, pp. 5336–5347, Dec. 2013.

[14] B. Zhang, B. V. K. V. Kumar, and D. Zhang, "Detecting diabetes mellitus and nonproliferative diabetic retinopathy using tongue color, texture, and geometry features," *IEEE Trans. Biomed. Eng.*, vol. 61, no. 2, pp. 491–501, Feb. 2014.

[15] J. Li, D. Zhang, Y. Li, J. Wu, and B. Zhang, "Joint similar and specific learning for diabetes mellitus and impaired glucose regulation detection," *Inf. Sci.*, vol. 384, pp. 191–204, Apr. 2017.

[16] R. Miotto, F. Wang, S. Wang, X. Jiang, and J. T. Dudley, "Deep learning for healthcare: Review, opportunities and challenges," *Briefings Bioinformat.*, vol. 19, no. 6, pp. 1236–1246, 2017.

[17] P. Rodriguez et al., "Deep pain: Exploiting long short-term memory networks for facial expression classification," *IEEE Trans. Cybern.*, to be published.

[18] H. Greenspan, B. Van Ginneken, and R. M. Summers, "Guest editorial deep learning in medical imaging: Overview and future promise of an exciting new technique," *IEEE Trans. Med. Imag.*, vol. 35, no. 5, pp. 1153–1159, May 2016.

[19] G. Litjens et al., "A survey on deep learning in medical image analysis," *Med. Image Anal.*, vol. 42, pp. 60–88, Dec. 2017.

[20] H. Chen et al., "Low-dose CT with a residual encoder–decoder convolutional neural network," *IEEE Trans. Med. Imag.*, vol. 36, no. 12, pp. 2524–2535, Dec. 2017.

[21] J. Zhang, M. Liu, and D. Shen, "Detecting anatomical landmarks from limited medical imaging data using two-stage task-oriented deep neural networks," *IEEE Trans. Image Process.*, vol. 26, no. 10, pp. 4753–4764, Oct. 2017.

[22] P. Yan, Y. Cao, Y. Yuan, B. Turkbey, and P. L. Choyke, "Label image constrained multiatlas selection," *IEEE Trans. Cybern.*, vol. 45, no. 6, pp. 1158–1168, Jun. 2015.

[23] L. Lin, W. Yang, C. Li, J. Tang, and X. Cao, "Inference with collaborative model for interactive tumor segmentation in medical image sequences," *IEEE Trans. Cybern.*, vol. 46, no. 12, pp. 2796–2809, Dec. 2016.

[24] A. Kumar, J. Kim, D. Lyndon, M. Fulham, and D. Feng, "An ensemble of fine-tuned convolutional neural networks for medical image classification," *IEEE J. Biomed. Health Informat.*, vol. 21, no. 1, pp. 31–40, Jan. 2017.

[25] H. C. Shin et al., "Deep convolutional neural networks for computer-aided detection: CNN architectures, dataset characteristics and transfer learning," *IEEE Trans. Med. Imag.*, vol. 35, no. 5, pp. 1285–1298, May 2016.

[26] Q. Tang, Y. Liu, and H. Liu, "Medical image classification via multiscale representation learning," *Artif. Intell. Med.*, vol. 79, pp. 71–78, Jun. 2017.

[27] Z. Zhang, Y. Xie, F. Xing, M. Mcgough, and L. Yang, "MDNet: A semantically and visually interpretable medical image diagnosis network," in *Proc. Int. Conf. Comput. Vis. Pattern Recognit.*, 2017, pp. 6428–6436.

[28] Y. Bao and X. Jiang, "An intelligent medicine recommender system framework," in *Proc. IEEE 11th Conf. Ind. Electron. Appl. (ICIEA)*, 2016, pp. 1383–1388.

[29] Y. Zou, P. X. Liu, Q. Cheng, P. Lai, and C. Li, "A new deformation model of biological tissue for surgery simulation," *IEEE Trans. Cybern.*, vol. 47, no. 11, pp. 3494–3503, Nov. 2017.

[30] B. Lei, P. Yang, T. Wang, S. Chen, and D. Ni, "Relational-regularized discriminative sparse learning for Alzheimer's disease diagnosis," *IEEE Trans. Cybern.*, vol. 47, no. 4, pp. 1102–1113, Apr. 2017.

[31] T. Hoang et al., "Detecting signals of detrimental prescribing cascades from social media," *Artif. Intell. Med.*, vol. 71, pp. 43–56, Jul. 2016.

[32] C. Feichtenhofer, A. Pinz, and A. P. Zisserman, "Convolutional two-stream network fusion for video action recognition," in *Proc. Comput. Vis. Recognit. Pattern*, 2016, pp. 1933–1941.

[33] A. Karpathy et al., "Large-scale video classification with convolutional neural networks," in *Proc. Comput. Vis. Recognit. Pattern*, 2014, pp. 1725–1732.

[34] J. Long and M. J. Yuan, "A novel clinical decision support algorithm for constructing complete medication histories," *Comput. Methods Programs Biomed.*, vol. 145, pp. 127–133, Jul. 2017.

[35] T. Yu et al., "Knowledge graph for TCM health preservation: Design, construction, and applications," *Artif. Intell. Med.*, vol. 77, pp. 48–52, Mar. 2017.

[36] F. Liao, X. Chen, X. Hu, and S. Song, "Estimation of the volume of the left ventricle from MRI images using deep neural networks," *IEEE Trans. Cybern.*, vol. 49, no. 2, pp. 495–504, Feb. 2017.

[37] A. Esteva et al., "Dermatologist-level classification of skin cancer with deep neural networks," *Nature*, vol. 542, no. 7639, pp. 115–118, 2017.

[38] L. Yao et al., "Discovering treatment pattern in traditional Chinese medicine clinical cases by exploiting supervised topic model and domain knowledge," *J. Biomed. Informat.*, vol. 58, pp. 260–267, Dec. 2015.

[39] K. Simonyan and A. Zisserman, "Very deep convolutional networks for large-scale image recognition," *arXiv preprint arXiv:1409.1556*, 2014.

[40] C. Szegedy et al., "Going deeper with convolutions," in *Proc. Comput. Vis. Pattern Recognit.*, 2015, pp. 1–9.

[41] K. He, X. Zhang, S. Ren, and J. Sun, "Deep residual learning for image recognition," in *Proc. Comput. Vis. Pattern Recognit.*, 2016, pp. 770–778.

[42] T. L. Griffiths and M. Steyvers, "Finding scientific topics," in *Proc. Nat. Acad. Sci. USA*, 2004, pp. 5228–5235.

[43] S. Kullback, *Information Theory and Statistics*, Courier Corporat., 1997.

[44] L. Breiman, "Random forests," *Mach. Learn.*, vol. 45, no. 1, pp. 5–32, 2001.

[45] J. Mairal, F. R. Bach, J. Ponce, and G. Sapiro, "Online dictionary learning for sparse coding," in *Proc. Int. Conf. Mach. Learn.*, 2009, pp. 689–696.

[46] S. S. Girija. (2016). *TensorFlow: Large-Scale Machine Learning on Heterogeneous Distributed Systems*. [Online]. Available: https://www.tensorflow.org/

[47] Y. Feng, Z. Wu, X. Zhou, Z. Zhou, and W. Fan, "Knowledge discovery in traditional Chinese medicine: State of the art and perspectives," *Artif. Intell. Med.*, vol. 38, no. 3, pp. 219–236, 2006.

[48] W. Li, Z. Yang, and X. Sun, "Exploration on generating traditional Chinese medicine prescription from symptoms with an end-to-end method," *arXiv preprint arXiv:1801.09030*, 2018.

[49] J. Deng et al., "ImageNet: A large-scale hierarchical image database," in *Proc. Comput. Vis Pattern Recognit.*, 2009, pp. 248–255.

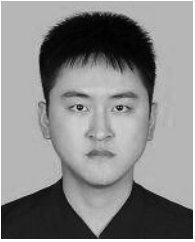

**Yang Hu** received the M.A.Eng. degree from the Kunming University of Science and Technology, Kunming, China, in 2016. He is currently pursuing the Ph.D. degree with the South China University of Technology, Guangzhou, China.

His current research interests include neural network and biomedical information processing.

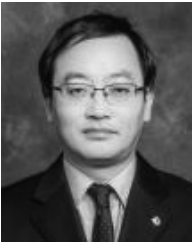

**Guihua Wen** received the Ph.D. degree in computer science and engineering from the South China University of Technology, Guangzhou, China.

He is currently a Professor and a Doctoral Supervisor with the School of Computer Science and Technology, South China University of Technology, where he is also the Professor in Chief with Data Mining and Machine Learning Laboratory and the Director of the Artificial Intelligence Chinese Medicine Engineering Research Institute. His current research interests include cognitive affective computing, machine learning, and data mining.

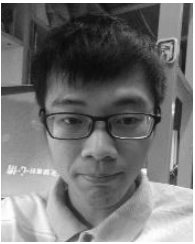

**Huiqiang Liao** is currently pursuing the master's degree with the College of Computer Science and Engineering, South China University of Technology, Guangzhou, China.

His current research interests include image processing and deep learning.

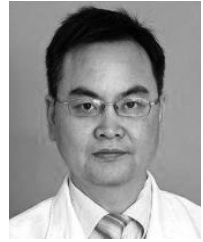

**Changjun Wang** received the Ph.D. degree from the Shanghai University of Traditional Chinese Medicine, Shanghai, China.

He is currently a Professor and a Doctoral Supervisor with the School of Medicine of South China University of Technology. He is also the Director with the Department of Traditional Chinese Medicine, Guangdong General Hospital, Guangzhou, China.

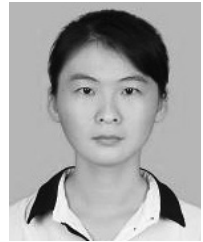

**Dan Dai** received the M.A.Eng. degree from the Kunming University of Science and Technology, Kunming, China, in 2016. She is currently pursuing the Ph.D. degree with the South China University of Technology, Guangzhou, China.

Her current research interests include machine learning and biomedical information processing.

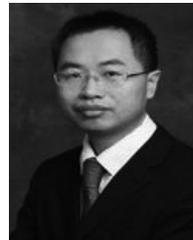

**Zhiwen Yu** (S'06–M'08–SM'14) received the Ph.D. degree from the City University of Hong Kong, Hong Kong, in 2008.

He is a Professor with the School of Computer Science and Engineering, South China University of Technology, Guangzhou, China, and an Adjunct Professor with Sun Yat-sen University, Guangzhou. He has published over 100 referred journal papers and international conference papers, including the IEEE TRANSACTIONS ON KNOWLEDGE AND DATA ENGINEERING, the IEEE TRANSACTIONS ON EVOLUTIONARY COMPUTATION, the IEEE TRANSACTIONS ON CYBERNETICS, the IEEE TRANSACTIONS ON MULTIMEDIA, the IEEE TRANSACTIONS ON CIRCUITS AND SYSTEMS FOR VIDEO TECHNOLOGY, the IEEE/ACM TRANSACTIONS ON COMPUTATIONAL BIOLOGY AND BIOINFORMATICS, the IEEE TRANSACTIONS ON NANOBIOSCIENCE, *Information Sciences*, *Pattern Recognition*, *Bioinformatics*, and SIGKDD. His current research interests include data mining, machine learning, bioinformatics, and pattern recognition.

Dr. Yu is a Senior Member of ACM, IRSS, China Computer Federation, and Chinese Association for Artificial Intelligence.